\documentclass[11pt,a4paper]{article}
\usepackage[hyperref]{acl2021}
\usepackage{times}
\usepackage{latexsym}

\usepackage{microtype}

\usepackage{epsfig}
\usepackage{graphicx}
\usepackage{subfigure}
\usepackage{overpic}
\usepackage{amsfonts}
\usepackage{booktabs}
\usepackage{bm}
\usepackage{amsmath}
\usepackage{bbm}
\usepackage{url}
\usepackage{color}

\newcommand{\figref}[1]{Figure~\ref{#1}}
\newcommand{\tabref}[1]{Table~\ref{#1}}

\usepackage{booktabs}
\usepackage{threeparttable}
\usepackage{multirow}

\aclfinalcopy 
\def\aclpaperid{232} 

\title{GoG: Relation-aware Graph-over-Graph Network for Visual Dialog}

\author{{Feilong Chen}, {Xiuyi Chen}, {Fandong Meng}, {Peng Li},  {Jie Zhou} \\
  Pattern Recognition Center, WeChat AI, Tencent Inc, Beijing, China \\
  {\tt
  \{\href{mailto:ivess.chan@gmail.com}{ivess.chan},\href{mailto:hugheren.chan@gmail.com
}{hugheren.chan}\}@gmail.com}\\
  {\tt \{\href{mailto:fandongmeng@tencent.com}{fandongmeng},\href{mailto:patrickpli@tencent.com}{patrickpli},\href{mailto:withtomzhou@tencent.com}{withtomzhou}\}@tencent.com}\\
  }

\date{}

\begin{document}
\maketitle
\begin{abstract}
    Visual dialog, which aims to hold a meaningful conversation with humans about a given image, is a challenging task that requires models to reason the complex dependencies among visual content, dialog history, and current questions. Graph neural networks are recently applied to model the implicit relations between objects in an image or dialog. However, they neglect the importance of 1) coreference relations among dialog history and dependency relations between words for the question representation; and 2) the representation of the image based on the fully represented question. Therefore, we propose a novel relation-aware graph-over-graph network (GoG) for visual dialog. Specifically, GoG consists of three sequential graphs: 1) H-Graph, which aims to capture coreference relations among dialog history; 2) History-aware Q-Graph, which aims to fully understand the question through capturing dependency relations between words based on coreference resolution on the dialog history; and 3) Question-aware I-Graph, which aims to capture the relations between objects in an image based on fully question representation. As an additional feature representation module, we add GoG to the existing visual dialogue model. Experimental results show that our model outperforms the strong baseline in both generative and discriminative settings by a significant margin.
\end{abstract}

\section{Introduction}

\begin{figure}[t!]
\centering
\scalebox{0.98}{
  \begin{overpic}[width=\columnwidth]{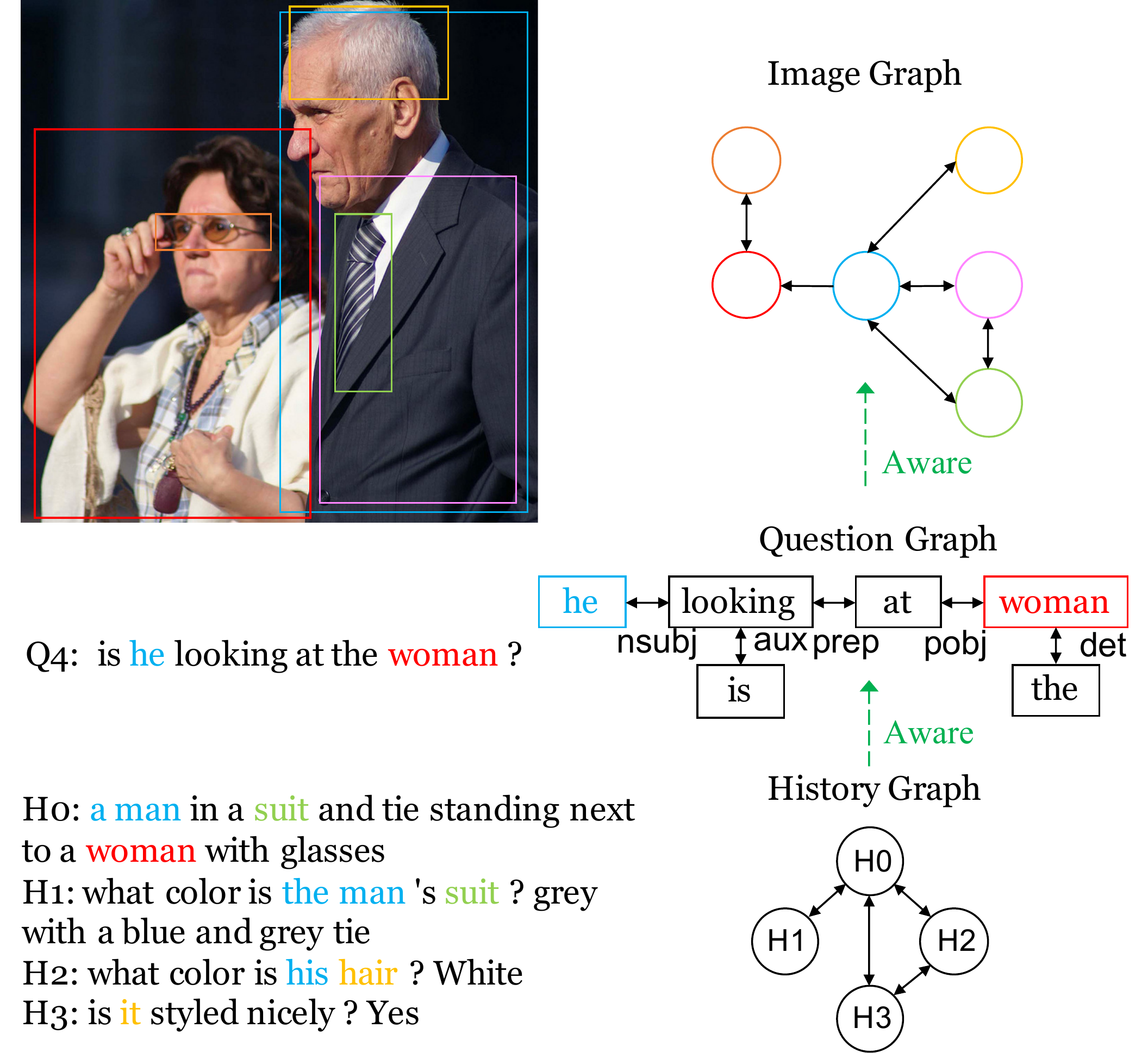}
  \end{overpic}
  }
  \caption{An example of complex relationships in visual dialog. The color in the text corresponds to the same color box in the image, which indicates the same entity. In visual dialog, we construct three graphs. History graph (H-Graph): arrows indicate the coreference relations between QA pairs in dialog history. Question graph (Q-Graph): arrows indicate dependency relations of the question. Image graph (I-Graph): arrows indicate spatial relations between objects in an image. The dark green dotted line indicates the bottom graph affects the upper graph.
  }\label{fig:relation} 
\end{figure}

Vision-language tasks have drawn more attention with the development of multi-modal natural language processing~\cite{baltruvsaitis2018multimodal,chen2020bridging,chen2019working}, such as image captioning~\cite{xu2015show,Anderson2016SPICE,anderson2018bottom,cornia2020m2,ghanimifard2019goes}, visual question answering~\cite{ren2015exploring,gao2015you,lu2016hierarchical,anderson2018bottom,li2019relation,huang2020aligned} and visual dialog~\cite{das2017visual,Kottur2018VisualCR,agarwal2020history,wang2020vd,Qi2020TwoCP,chen2021learning}. Relations in these tasks are significant for reasoning and understanding the textual and visual information. Specifically, visual dialog, which aims to hold a meaningful conversation with a human about a given image, is a challenging task that requires models to reason complex relations among visual content, dialog history, and current questions.

Kinds of attention mechanisms are served as the backbone of previous mainstream approaches~\cite{lu2017best,wu2018you,Kottur2018VisualCR,gan2019multi,guo2019dual}, following~\citeauthor{das2017visual}~\citeyear{das2017visual}. HCAIE~\cite{lu2017best} provides a history-conditioned image attentive encoder to represent the question, the question-attended history, and the attended image. CoAtt~\cite{wu2018you} provides a sequential co-attention encoder to realize that each input feature is co-attended by the other two features in a sequential fashion. ReDAN~\cite{gan2019multi} and DMAM~\cite{chen2020dmrm} use multi-step reasoning based on dual attention to answer a series of questions about an image. DAN~\cite{guo2019dual}, MCAN~\cite{agarwal2020history} and LTMI~\cite{nguyenefficient} utilize multi-head attention mechanisms to manage multi-modal intersection. However, these approaches tend to catch only the most discriminative information, ignoring other rich complementary clues, such as relations between objects in an image. 

\begin{figure*}[t]
\centering
\scalebox{0.99}{
  \begin{overpic}[width=\textwidth]{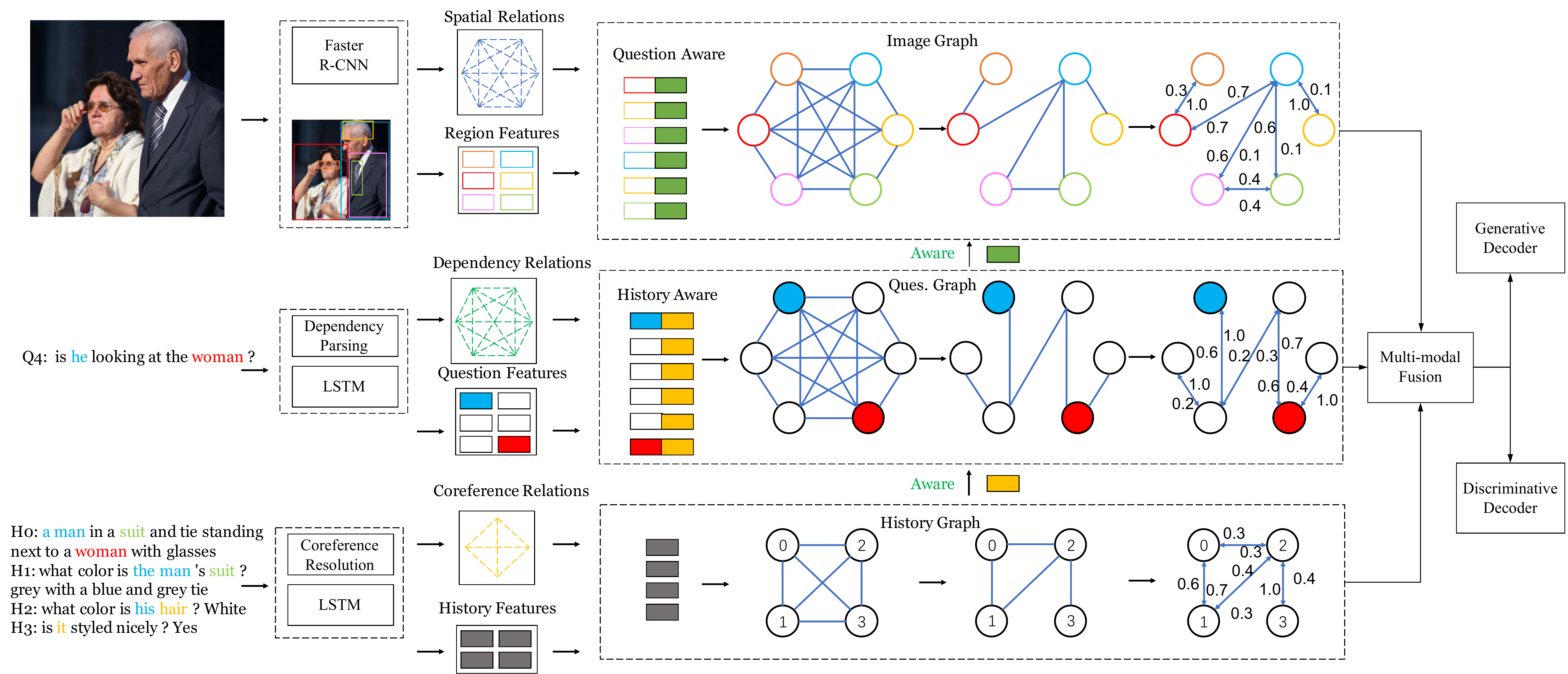}
  \end{overpic}
  }
  \caption{Framework of our Relation-aware Graph-over-Graph Network.
  }\label{fig:model} 
\end{figure*}

Recent visual dialog studies~\cite{zheng2019reasoning,schwartz2019factor,jiang2020dualvd,guo2020iterative,jiang2020kbgn} explore the higher-level semantic representation of images or dialog history, notably with graph-based structures for modeling the image or dialog history. 
Although graph-based structures have been considered, these graph-based models lack explicitly capturing complex relations within visual content or textual contexts, and relations between them. As shown in~\figref{fig:relation}, there are complex relations such as coreference relations among dialog history, dependency relations between words in the question, spatial relations between objects in the image. For example, to answer the question Q4 {\em ``is he looking at the woman ?''}, we firstly need to reason in dialog history to know who {\em ``he''} is, then further understand the intention of the question with the understood of history and syntax of questions, and finally know clearly the spatial location and relation about {\em ``the man''} and {\em ``the woman''} in the image based on fully question understanding. How to 1) understand the coreference among history, 2) understand the intention of the question with its syntax and history, 3) understand the image with fully question understanding are worth exploring.

Therefore, in this paper, we propose a novel relation-aware graph-over-graph network (GoG) for visual dialog. Specifically, GoG consists of three sequential graphs: 1) H-Graph, which aims to capture coreference relations among dialog history; 2) History-aware Q-Graph, which aims to fully represent the question through capturing dependency relations between words based on coreference resolution on the dialog history; and 3) Question-aware I-Graph, which aims to capture the relations between objects in an image on the basis of fully question representation. As an additional feature representation module, we add GoG to the strong visual dialogue model LTMI~\cite{nguyenefficient}. We test the effectiveness of our proposed model on two large-scale datasets: VisDial v0.9 and v1.0~\cite{das2017visual}. Both automatic and manual evaluations show that our approach can be used to improve the prior strong models. The contributions of this work are summarized as follows:
\begin{itemize}
  \item We explore how to construct complex explicit relations in visual dialog, i.e., coreference relations among dialog history, dependency relations between words in the question, spatial relations between objects in the image.
  \item We propose a novel relation-aware graph-over-graph network to reason relations within and among different graphs to obtain a high-level representation of multi-modal information and use it to generate a visually and contextually coherent response.
  \item We conduct extensive experiments and ablation studies on two large-scale datasets VisDial v0.9 and v1.0. Experimental results show that our GoG model can be used to improve the previous strong visual dialog model in both generative and discriminative settings.
\end{itemize}

\section{Relation-aware Graph-over-Graph Network}
\subsection{Preliminary}
Following Das et al.~\cite{das2017visual}, a visual dialog agent is given three inputs, i.e., an image $I$, dialog history (the caption and question-answer pairs) till round $t-1$: $H=(\underbrace{Cap}_{H_0}, \underbrace{(Q_1, A_1)}_{H_1}, \cdots, \underbrace{(Q_{t-1},A_{t-1})}_{H_{t-1}})$ and the current question $Q_t$ at round $t$, where $Cap$ is the caption describing the image taken as $H_0$, and $H_1, \dots, H_{t-1}$ are concatenations of question-answer pairs. The goal of the visual dialog agent is to generate a response $A_t$ to the question $Q_t$.

As shown in~\figref{fig:model}, our relation-aware graph-over-graph network (GoG) firstly takes the image, the dialog history, and the question as inputs and represent them using Faster RCNN~\cite{ren2015faster} and LSTM~\cite{hochreiter1997long}. Secondly, GoG constructs the history graph, the history-aware question graph, and the question-aware image graph. Thirdly, GoG utilizes the attention alignment module to fuse the three graphs. Finally, GoG uses the fused multi-modal information to give corresponding answers.

Firstly, we simply describe the feature representation of three inputs. Secondly, we introduce our graph attention. Then we describe how we apply our graph attention to the history graph, question graph and image graph to construct our graph-over-graph network. Finally, we describe how we apply our graph-over-graph network to the strong visual dialog models.

\subsection{Feature Representation}
Similar to~\cite{anderson2018bottom}, we extract the
image features by using a pretrained Faster RCNN~\cite{ren2015faster}. We select $\mu$ object proposals for each image, where each object proposal is represented by a 2048-dimension feature vector. The obtained visual region features are denoted as $v = {v_{i=0}^{\mu}} \in R^{\mu \times d_v}$.

To extract the question features, each word is embedded into a 300-dimensional vector initialed with the Glove vector~\cite{pennington2014glove}. The word embeddings are taken as inputs by an LSTM encoder~\cite{hochreiter1997long}, which produces the initial question representation $q \in R^{\lambda \times d_q}$. Each history sentence features are obtained as same as the question features. We concatenate the last state $h^{last} \in R^{d_h}$ of each turn history features to get the initial history representation $h = {h_{0}^{t-1}} = [h_{0}^{last}, \dots, h_{t-1}^{last}] \in R^{t \times d_h}$. $\lambda$ denotes the length of the question, $t$ denotes the turn of dialog history, $d_q$ denotes the dimension of each word in questions, $d_h$ denotes the dimension of each word in history and $[\cdot, \cdot]$ denotes the concatenation operation.

\subsection{Graph Attention}
Given a target node $i$ and a neighboring node $j \in \mathcal{N}(i)$ with a $k \times k$ adjacency matrix $R$, where $\mathcal{N}(i)$ is the set of $k$ nodes neighboring with node $i$, and the representations of node $i$ and node $j$ are $u_i$ and $u_j$, respectively. To obtain the correlation score $s_{ij}$ between node $i$ and $j$, self-attention~\cite{vaswani2017attention} is then performed on the vertices, which generates a relation score $s_{ij}$ between node features $u_i$ and $u_j$:
\begin{equation}
    s_{ij} = \frac{(U_iu_i)^T \cdot V_ju_j}{\sqrt{d_u}}, \label{u_begin}
\end{equation}
where $U_i$ and $V_j$ are trainable parameters. We apply a softmax function over the correlation score $s_{ij}$ to obtain weight $\alpha_{ij}$ :
\begin{equation}
    \alpha_{ij} = \frac{{\rm exp}(s_{ij} + c_{u, lab(i,j)})}{\sum_{j \in \mathcal{N}(i)}{\rm exp}(s_{ij} + c_{u, lab(i,j)})},
\end{equation}
where $c_{\{\cdot \}} = W_{lab}A_{ij}$ is a bias term, $lab(i,j)$ represents the label of each edge, and $W_{lab}$ is a learned parameter. The representations of neighboring nodes $u_j$ are first transformed via a learned linear transformation with $W_u$. Those transformed representations are then gathered with weight $\alpha_{ij}$ , followed by a non-linear function $\sigma$. This propagation can be denoted as:
\begin{equation}
    u^{*}_i = \sigma \Big ( u_i + \sum_{j \in \mathcal{N}(i)}R_{ij}\alpha_{ij}W_uu_j \Big ). \label{u_end}
\end{equation}
We utilize ${\rm GraphAtt}(\cdot)$ to denote equations from Eq.~\eqref{u_begin} to Eq.~\eqref{u_end}

\subsection{History Graph Construction}
\begin{figure}[t!]
\centering
\scalebox{0.98}{
  \begin{overpic}[width=\columnwidth]{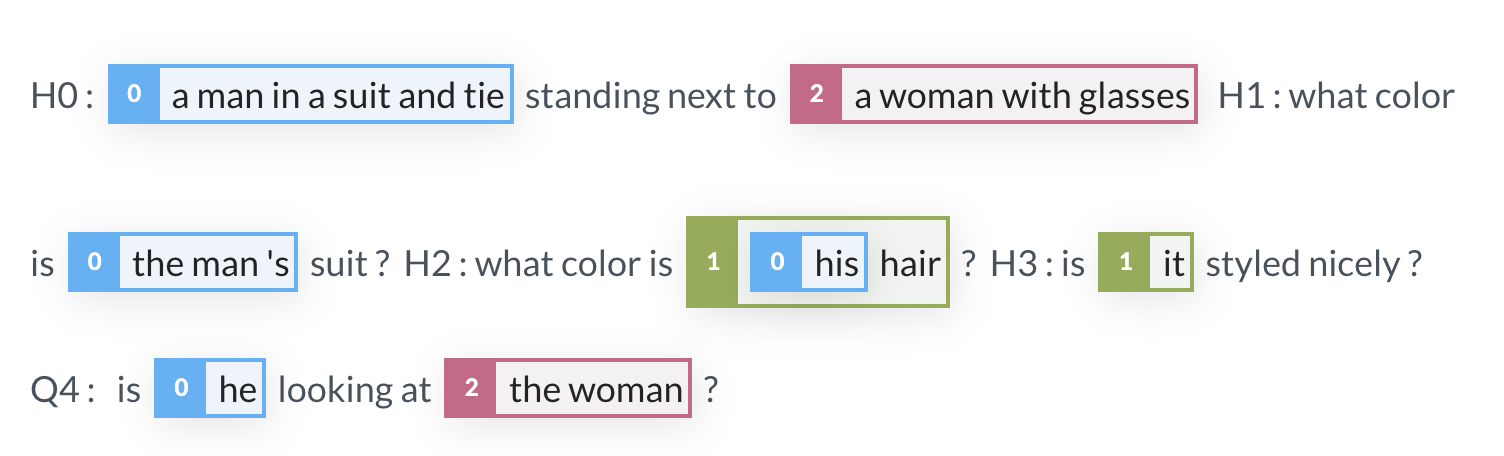}
  \end{overpic}
  }
  \caption{The history is performed by coreference resolution. The same color box and the same number indicate the coreference relation between different expressions of the same entity.
  }\label{fig:hisrelation} 
\end{figure}

In practice, we observe that coreference relations exist in dialog history. To fully understand the coreference among dialog history, we utilize the coreference resolution tool~\cite{lee2017end} to identify coreference relations. We use the caption and questions to identify the relations instead of QA pairs because there is no ground truth answer in the test split. As shown in~\figref{fig:hisrelation}, we provide a four-turn dialog to show coreference resolution. The same color boxes with the same numbers indicate the coreference relations. For example, the blue box with number 0 indicates they are related to the word ``a man'' with its attribute ``in a suit and tie''.

 
\paragraph{Pruned History Graph with Coreference Relations.}
We treat each turn history as a node. By analyzing the coreference relations of the history, we obtain the relations between history as shown in~\figref{fig:hisrelation}. According to coreference relations, we construct a sparse graph, as shown in the history graph of~\figref{fig:model}.

\paragraph{History Graph Attention.}
Given a graph with $t$ nodes, i.e. a $t$-turn dialog, each turn representation in history is a node. We represent the graph structure with a $t \times t$ adjacency matrix $A$, where $A_{ij} = 1$ if there is a coreference relation between node $i$ and node $j$; else $A_{ij} = 0$. 

The relation-aware graph based history representation $h^{*}_i$ is as follows:
\begin{equation}
    h^{*}_i = {\rm GraphAtt}(h_i, A)
\end{equation}

\begin{figure}[t!]
\centering
\scalebox{0.90}{
  \begin{overpic}[width=\columnwidth]{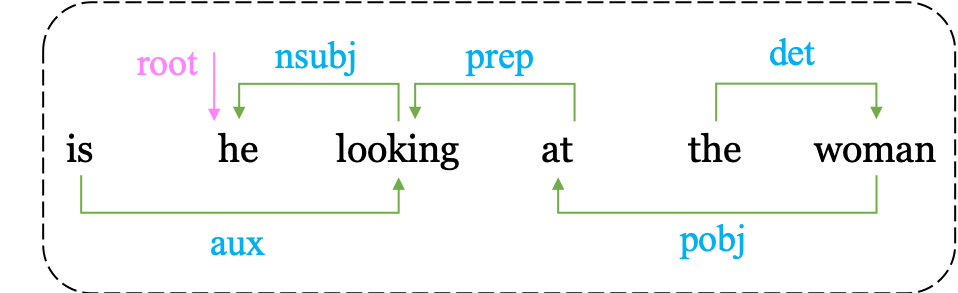}
  \end{overpic}
  }
  \caption{The question is performed by dependency parsing. The word in pink is the root node. The direction of green arrows indicates the dependency relation between two words, and the blue words (e.g., det, dobj) are relation types.
  }\label{fig:quesrelation} 
\end{figure}

\subsection{Question Graph Construction}
In practice, we observe that two words in a sentence usually hold a certain relation. Such relations can be identified by the Neural Dependency Parsing~\cite{dozat2017deep}. 
 

\paragraph{Pruned Question Graph with Dependency Relations.}
We treat each word in a question as a vertex. By parsing the dependency relations of a question, we obtain the relations between words as shown in~\figref{fig:quesrelation}. According to dependency relations, we obtain our sparse question graph, as shown in the question graph of~\figref{fig:model}. 

\paragraph{History-aware Question Graph Attention.}
Given a graph with $\lambda$ nodes, each word in a question is a node. We represent the graph structure with a $\lambda \times \lambda$ adjacency matrix $B$, where $B_{ij} = 1$ if there is a dependency relation between node $i$ and node $j$; else $B_{ij} = 0$. 

In order to utilize the history to help understand questions, we use a history-aware attention mechanism to inject semantic information from the history into the question graph. The aware history representation is calculated as follows:
\begin{equation}
    \hat{h} ={\rm softmax}\big (W_{h_1}\sigma(W_{h_2}h^{*}) \big )h^{*},
\end{equation}
where $\hat{h} \in R^{d_h}$, $h^{*}$ is the final representation of dialog history. $W_{h_1}$ and $W_{h_2}$ are learned parameters. The history-aware question features are achieved by concatenating the adaptive history representation $\hat{h}$ with each of question features $q_i$, denoted as:
\begin{equation}
    q^{\prime}_i = [q_i, \hat{h}], \quad {\rm for}\quad i=1,\dots,\lambda.
\end{equation}

The history-aware and relation-aware graph based question representation $q^{*}_i$ is as follows:
\begin{equation}
    q^{*}_i = {\rm GraphAtt}(q^{\prime}_i, B)
\end{equation}

\subsection{Image Graph Construction}
 
\paragraph{Pruned Image Graph with Spatial Relations.}
By treating each object region in an image as a vertex, we can construct a fully-connected undirected graph, as shown in the image graph of~\figref{fig:model}. Each edge represents a relation between two object regions.
Spatial relations represent an object position in an image, which correspond to a 4-dimension spatial coordinate $[x_1, y_1, x_2, y_2]$. Note that $(x_1, y_1)$ is the coordinate of the top-left point of the bounding box and $(x_2, y_2)$ is the coordinate of the bottom-right point of the bounding box. Following Yao et al.~\cite{yao2018exploring}, we classify different spatial relations into 11 different categories, such as {\em inside}, {\em cover} and {\em overlap}. We utilize the overlapping region between two object regions to judge whether there is an edge between two regions. If two object regions have overlapping parts, it means that there is a strong correlation between these two objects. If two object regions are too far away from each other, it means that there is no relation between these two objects. According to the spatial relations, we prune some irrelevant relations between objects and obtain a sparse graph, as shown in the image graph of~\figref{fig:model}.
 
\paragraph{Question-aware Image Graph Attention.}
Given a graph with $\mu$ nodes, each object in an image is a node. We represent the graph structure with a $\mu \times \mu$ adjacency matrix $D$, where $D_{ij} = 1$ if there is a spatial relation between node $i$ and node $j$; else $D_{ij} = 0$. 

Based on the fully question understanding, we use a question-aware attention mechanism to inject semantic information from the question into the image graph. The adaptive question representation is calculated as follows:
\begin{equation}
    \hat{q} ={\rm softmax}\big (W_{q_1}\sigma(W_{q_2}q^{*}) \big )q^{*},
\end{equation}
where $W_{q_1}$ and $W_{q_2}$ are learned parameters.
The question-aware image features are achieved by concatenating the aware question representation $\hat{q}$ with each of the $\mu$ image features $v_i$, denoted as:
\begin{equation}
    v^{\prime}_i = [v_i, \hat{q}], \quad {\rm for}\quad i=1,\dots,\mu.
\end{equation}

The question-aware and relation-aware graph based image representation $v^{*}_i$ is as follows:
\begin{equation}
    v^{*}_i = {\rm GraphAtt}(v^{\prime}_i, D)
\end{equation}

\subsection{Multi-modal Fusion}
After obtaining the relation-aware representation, we fuse the question representation $q^{*}$, history representation $h^{*}$, visual representation $v^{*}$ through a multi-modal fusion strategy. We can use any existing visual dialog models to learn a joint representation $\mathcal{J}$:
\begin{equation}
    \mathcal{J} = \mathcal{F}(q^{*},h^{*},v^{*};\Theta),
\end{equation}
where $\mathcal{J}$ is a visual dialog model and $\Theta$ are trainable parameters of the fusion module. The design of generative and discriminative decoders~\cite{das2017visual}, and multi-task learning strategy~\cite{nguyenefficient} can be referred to Appendix~{\ref{appendix:model}}.

\begin{table*}
\centering
\resizebox{0.97 \textwidth}!{
\begin{tabular}{l|ccccc|cccccc}
\toprule
\multirow{2}{*}{Model} &
\multicolumn{5}{c|}{VisDial v0.9 (val)} &
\multicolumn{6}{c}{VisDial v1.0 (val)} \\
\cline{2-6}   \cline{7-12}
& MRR $\uparrow$ & R@1 $\uparrow$ & R@5 $\uparrow$ & R@10 $\uparrow$ & Mean $\downarrow$ & NDCG $\uparrow$ & MRR $\uparrow$ & R@1 $\uparrow$ & R@5 $\uparrow$ & R@10 $\uparrow$ & Mean $\downarrow$  \\
\midrule

\multicolumn{12}{c}{Attention-based Model} \\
\midrule
RvA~\cite{niu2019recursive} & 55.43 & 45.37 & 65.27 & \underline{72.97} & \underline{10.71}
& - & - & - & - & - & - \\
DVAN~\cite{guo2019dual} & 55.94 & \underline{46.58} & 65.50 & 71.25 & 14.79
& - & - & - & - & - & - \\
DMRM~\cite{chen2020dmrm} & \underline{55.96} & 46.20 & \underline{66.02} & 72.43 & 13.15
& - & 50.16 & 40.15 & 60.02 & 67.21 & 15.19 \\
DAM~\cite{jiang2020dam} & - & - & - & - & -
& 60.93 & \underline{50.51} & \underline{40.53} & \underline{60.84} & 67.94 & 16.65 \\
\midrule
\multicolumn{12}{c}{Pretraining-based Model} \\
\midrule
VDBERT~\cite{wang2020vd}$^\diamond$ & 55.95 & 46.83 & 65.43 & 72.05 & 13.18
& - & - & - & - & - & - \\
\midrule
\multicolumn{12}{c}{Graph-based Model} \\
\midrule
KBGN~\cite{jiang2020kbgn} & - & - & - & - & -
& 60.42 & 50.05 & 40.40 & 60.11 & 66.82 & 17.54 \\
\midrule
LTMI~\cite{nguyenefficient}$^\dagger$ & 55.85 & 46.07 & 65.97 & 72.44 & 14.17
& \underline{61.61} & 50.38 & 40.30 & 60.72 & \underline{68.44} & \underline{15.73} \\
LTMI-GoG (Ours) & \bf{56.32} & \bf{46.65} & \bf{66.41} & 72.69 & 13.78
& \bf{62.63} & \bf{51.32} & \bf{41.25} & \bf{61.83} & \bf{69.44} & \bf{15.32} \\
LTMI-GoG-Multi (Ours) & 56.89 & 47.04 & 66.92 & 72.87 & 13.45
& 63.35 & 51.80 & 41.78 & 62.23 & 69.79 & 15.16 \\
LTMI-GoG-Multi* (Ours) & 59.38& 48.58 & 71.33 & 78.78 & 9.94
& 65.20 & 55.38 & 43.93 & 68.22 & 76.75 & 9.98 \\
\bottomrule
\end{tabular}} 
\caption{Main comparisons on both VisDial v0.9 and v1.0 datasets using the generative decoder. $\dagger$ denotes that we re-implemented the model. $\diamond$ denotes that the model utilizes large extra datasets for training. Underline indicates the highest performance among previous approaches except pretraining-base models. (t-test, p-value$ \textless$0.01)} 
\label{tab:gen}
\end{table*}

\section{Experiments}
\subsection{Experiment Setup}
\paragraph{Datasets and Implementation Details.}

We conduct experiments on the VisDial v0.9 and v1.0 datasets~\cite{das2017visual} to verify our approach. VisDial v0.9 contains 83k dialogs on COCO-train~\cite{lu2017best} and 40k dialogs on COCO-val images as the test set, for a total of 1.23M dialog question-answer pairs. VisDial v1.0 dataset is an extension of VisDial v0.9 dataset with additional 10k COCO-like images. VisDial v1.0 dataset contains 123k, 2k, and 8k images as train, validation, and test splits, respectively.

To represent image regions, we use Faster R-CNN~\cite{ren2015faster} with ResNet-101~\cite{he2016deep} finetuned on the Visual Genome dataset~\cite{krishna2017visual}, thus obtaining a 2048-dimension feature vector for each region. Following ~\cite{nguyenefficient}, we detect K = 100 objects from each image. For the question and history features, we first build the vocabulary composed of 11,322 words that appear at least five times in the training split. The captions, questions, and answers are truncated or padded to 40, 20, and 20 words, respectively. We employ multi-head attention with 4 heads for all three graph attention networks. The dimension of hidden features is set to 512. 

Our model is implemented based on PyTorch~\cite{paszke2017automatic}. In experiments, we use Adam~\cite{kingma2014adam} optimizer for training, with the mini-batch size as 32. For the choice of the learning rate, we employ the warm-up strategy~\cite{goyal2017accurate}. Specifically, we begin with a learning rate of 0.0001, linearly increasing it at each epoch till it reaches 0.0002 at epoch 4. After 15 epochs, the learning rate is decreased by 1/4 for every 2 epochs up to 20 epochs. We use 4 Titan-XP GPU for training. We spend about 4 hour / 1 epoch for the discriminative setting and 1 hour / 1 epoch for the generative setting. The total parameter of our GoG model is 46.94M, while the total parameter of LTMI~\cite{das2017visual} is 42.20M. GoG only has an increase of 4.74M than LTMI.

\paragraph{Automatic Evaluation.}
We use a retrieval setting to evaluate individual responses at each round of a dialog, following~\cite{das2017visual}. Specifically, at test time, a list of 100-candidate answers is also given. The model is evaluated on retrieval metrics: (1) Rank of human response, (2) existence of the human response in $top-k$ ranked responses, i.e., R@$k$ (3) Mean reciprocal rank (MRR) of the human response and (4) Normalized discounted cumulative gain (NDCG) for VisDial v1.0.  

\paragraph{Human Evaluation.}
We randomly extract 100 samples for human evaluation~\cite{wu2018you} and then ask 3 human subjects to guess whether the last response in the dialog is human-generated or machine-generated. If at least 2 of them agree it is generated by a human, we think it passes the Truing Test (M1). We record the percentage of responses that are evaluated better than or equal to human responses (M2), according to the human subjects’ evaluation.

\begin{table*}
\centering
\resizebox{0.92\textwidth}!{
\begin{tabular}{l|ccccc|cccccc}
\toprule
\multirow{2}{*}{Model} &
\multicolumn{5}{c}{VisDial v0.9 (val)} &
\multicolumn{6}{c}{VisDial v1.0 (test-std)} \\
\cline{2-6}   \cline{7-12}
& MRR $\uparrow$ & R@1 $\uparrow$ & R@5 $\uparrow$ & R@10 $\uparrow$ & Mean $\downarrow$ & NDCG $\uparrow$ & MRR $\uparrow$ & R@1 $\uparrow$ & R@5 $\uparrow$ & R@10 $\uparrow$ & Mean $\downarrow$  \\
\midrule
\multicolumn{12}{c}{Attention-based Model} \\
\midrule
ReDAN~\cite{gan2019multi} & - & - & - & - & -
& 57.63 & 64.75 & 51.10  & 81.73  & 90.90  &  3.89  \\
MCA~\cite{agarwal2020history} & - & - & - & - & -
& \underline{72.73} & 37.68 & 20.67 & 56.67 & 72.12 & 8.89 \\
\midrule
\multicolumn{12}{c}{Pretraining-based Model} \\
\midrule
VisualBERT~\cite{murahari2019large}$^\diamond$ & - & - & - & - & - 
& 74.47 & 50.74 & 37.95 & 64.13 & 80.00 & 6.28 \\
VDBERT~\cite{wang2020vd}$^\diamond$ & 70.04 & 57.79 & 85.34 & 92.68 & 4.04
& 75.35 & 51.17 & 38.90 & 62.82 & 77.98 & 6.69 \\
\midrule
\multicolumn{12}{c}{Graph-based Model} \\
\midrule
GNN-EM~\cite{zheng2019reasoning} & 62.85 & 48.95 & 79.65 & 88.36 & 4.57
& 52.82 & 61.37 & 47.33 & 77.98 & 87.83 & 4.57 \\
DualVD~\cite{jiang2020dualvd} & 62.94 & 48.64 & 80.89 & 89.94 & 4.17
& 56.32 & 63.23 & 49.25 & 80.23 & 89.70 & 4.11 \\
FGA~\cite{schwartz2019factor} & 65.25 & 51.43 & 82.08 & 89.56 & 4.35
& 56.90 & \underline{66.20} & \underline{52.75} &  \underline{82.92} & \underline{91.07} & \underline{3.80} \\
CAG~\cite{guo2020iterative} & \underline{67.56} & \underline{54.64} & \underline{83.72} & \underline{91.48} & \underline{3.75} 
& 56.64 & 63.49 & 49.85 & 80.63 & 90.15 & 4.11 \\
KBGN~\cite{jiang2020kbgn} & - & - & - & - & -
& 57.60 &  64.13 &  50.47 & 80.70 &  90.16 &  4.08 \\
\midrule
LTMI~\cite{nguyenefficient}$^\dagger$ & 66.41 & 53.36 & 82.53 & 90.54 & 4.03
& 60.74 & 61.20 & 47.08 & 77.78 & 87.60 & 4.88 \\
LTMI-GoG (Ours) & 66.76 & 53.84 & 82.89 & 90.90 & 3.91
& 60.38 & 63.13 & 49.88 & 79.65 & 89.05 & 4.39 \\
LTMI-GoG-Multi (Ours) & 66.97 & 54.03 & 83.10 & 91.22 & 3.83
& 61.04 & 63.52 & 50.01 & 80.13 & 89.28 & 4.31  \\
\bottomrule
\end{tabular}} 
\caption{Main comparisons on both VisDial v0.9 and v1.0 datasets using the discriminative decoder. $\diamond$ denotes that the model utilizes large extra datasets for training. Underline indicates the highest performance among previous approaches except the pretraining-based models. (t-test, p-value$ \textless$0.01)} 
\label{tab:disc_test}
\end{table*}

\begin{table}[t]
    \centering
    \resizebox{0.98\columnwidth}!{
    \begin{tabular}{l|cccccc}
    \toprule
     Model & NDCG & MRR & R@1 & R@5 & R@10 & Mean \\
     \midrule
     ReDAN~\cite{gan2019multi} & -  & 64.29 & 50.65 &  81.29 & 90.17 & 4.10 \\
     KBGN & \underline{59.08} & \underline{64.86} & \underline{51.37} & \underline{81.71} & \underline{90.54} & \underline{4.00}  \\
     VDBERT~\cite{wang2020vd}$^\ddag$ & 56.20 & 62.25 & 48.16 & 79.57 & 89.01 & 4.31 \\
     VDBERT~\cite{wang2020vd}$^\diamond$ & 63.22 & 67.44 & 54.02 & 83.96 & 92.33 & 3.53 \\
     \midrule
     LTMI~\cite{nguyenefficient}$^\dagger$& 61.52  & 62.31 & 48.92 & 78.55 & 87.77 & 4.86 \\
     LTMI-GoG & \bf{62.24}  & 63.81 & 50.33 &  80.48 & 89.24 & 4.35 \\
     LTMI-GoG-Multi & 63.15 & 62.68 & 49.46 & 78.77 & 87.87 & 4.81 \\
    \bottomrule
    \end{tabular}}  
    \caption{Main comparisons on VisDial v1.0 val datasets using the discriminative decoder.} 
    \label{tab:disc_val}
\end{table}

\subsection{Main Results}
\paragraph{Baseline methods.}
In our experiment, compared methods can be grouped into four types: (1) Fusion-based models. (2) Attention-based models: ReDAN, CorefNMN, RvA, DVAN, DMRM, DAM. (3) The pretraining model: VDBERT and VisualBERT. (4) Graph-based models: GNN-EM, DualVD, FGA,
KBGN. Please refer to Appendix~{\ref{appendix:exp}} for more compared methods. 

GoG denotes our relation-aware graph-over-graph network. We use the strong model LTMI~\cite{nguyenefficient}\footnote{We reproduce results of LTMI by their official GitHub repo (https://github.com/davidnvq/visdial). We apply the default hyper-parameters as them.} as our multi-modal fusion module. LTMI is a very strong model which achieves some the-state-of-the-art results. ``Multi'' indicates the model uses multi-task learning at training but utilizes the generative or discriminative decoder at inference, respectively. ``Multi*'' indicates the model uses multi-task learning and utilizes the discriminative decoder to improve the generative decoder.  In general, our model outperforms the strong baseline by a significant margin. We use t-test to analyze our model and LTMI~\cite{nguyenefficient}. The p-values is less than 0.01, indicating that the results are significantly different.

\paragraph{Generative Results}
As shown in the right half of~\tabref{tab:gen}, we compare generative performance on the val v1.0 split. Our method improves significantly (about 1\% on all metrics) on the strong baseline LTMI~\cite{nguyenefficient} and outperforms all the compared methods on all metrics with large margins, which proves that GoG can improve the performance of visual dialog models by introducing explicit relation reasoning. Compared with the graph-based model KBGN~\cite{jiang2020kbgn}, our GoG-gen improves NDCG from 60.42 to 62.63 (+2.21\%), MMR from 50.05 to 51.32 (+1.27\%), which illustrates that our explicit relation reasoning is more effective because our approach reduce the noise of implicit relation modeling. LTMI-GoG-Multi and LTMI-GoG-Multi$^*$ obtain higher performance with large margins comparing with LTMI~\cite{nguyenefficient}, which shows that our approach is effective on multi-task setting. As shown in the left half of~\tabref{tab:gen}, we come to a similar conclusion on the val v0.9 split.  Our method improves a big margin (about 0.5\% on all metrics) on LTMI~\cite{nguyenefficient} and outperforms all the none pre-trained methods on MRR, R@1, and R@5.



\paragraph{Discriminative Results}
As shown in the right half of~\tabref{tab:disc_test}, our method improves a lot (near 1.5\% on all metrics except NDCG) based on LTMI~\cite{nguyenefficient} on the test-std v1.0 split. We also compare the performance on the val v1.0 split as shown in~\tabref{tab:disc_val}.  As shown in the left half of~\tabref{tab:disc_test}, we compare discriminative performance on the val v0.9 split. Our method improves a lot based on the LTMI~\cite{nguyenefficient}. As shown in~\tabref{tab:disc_val}, our approach outperforms VDBERT~\cite{wang2020vd}$^\ddag$ which trains from scratch without extra datasets.  All the comparison show that our approach is valid due to explicit relation modeling.

\begin{table}[tbp]
    \centering
    \resizebox{0.65\columnwidth}!{
    \begin{tabular}{c|l|c}
    \toprule
     Row & Model & NDCG \\
     \midrule
     & LTMI & 61.61 \\
     0 & LTMI-GoG & 62.63 \\
     \midrule
     1 & w/o I-Graph & 61.96 \\
     2 & w/o Q-Graph & 62.15 \\
     3 & w/o H-Graph & 62.03 \\
     \midrule
     4 & w/o Q-Aware & 62.41 \\
     5 & w/o H-Aware & 62.31\\
     \midrule
     6 & w/o Spatial Relation & 62.15 \\
     7 & w/o Dependency Relation & 62.24 \\
     8 & w/o Coreference Relation & 62.31 \\
    \bottomrule
    \end{tabular}}\vspace{3pt}
    \caption{Ablation study on VisDial v1.0 val datasets using the generative decoder.}\vspace{-13pt}
    \label{tab:ablation}
\end{table}

\subsection{Ablation Study}
As shown in~\tabref{tab:ablation}, we firstly remove the I-Graph, Q-Graph, H-Graph to validate the effect of each graph, respectively. Secondly, we validate the importance of concatenating operation. Finally, we use full connections to replace the relation in the graph to validate the importance of each relation. Firstly, the comparison between line 0 and line 1/2/3 shows all three graphs are crucial for visual dialog, leading to higher performance, and the I-Graph is most important. Secondly, the comparison between line 0 and line 4/5 shows that adding adaptive features gives a gain of approximately about +0.2. Thirdly, the comparison between line 1 and line 6/7/8 shows that doing graphs with relations gives better gain than simple fully-connected graphs. Spatial relation is the pick of the bunch because the full connection of 100 objects in an image will bring lots of noise. 



\begin{figure*}[t!]
\centering
\scalebox{0.95}{
  \begin{overpic}[width=\textwidth]{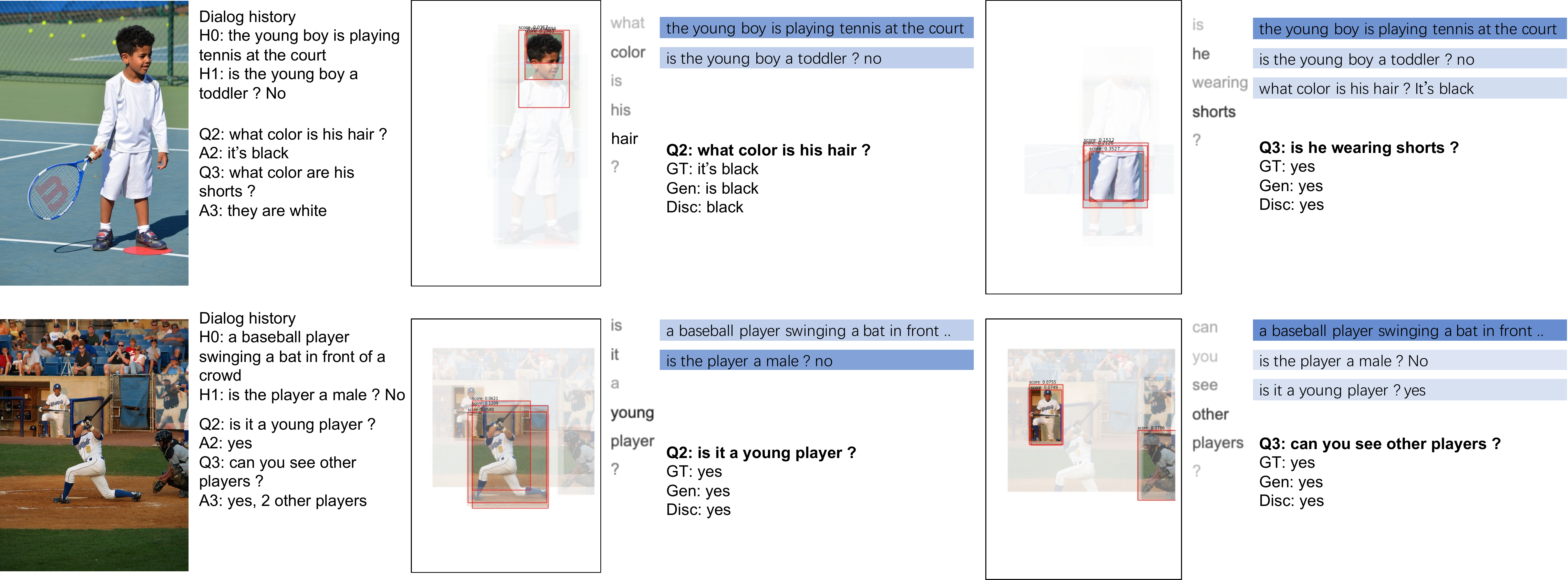}
  \end{overpic}
  }
  \caption{Visualization of attention maps generated in our model at two Q\&A rounds on two images.
  }\label{fig:example} 
\end{figure*}

\begin{figure*}[t]
\centering
\scalebox{0.90}{
  \begin{overpic}[width=\textwidth]{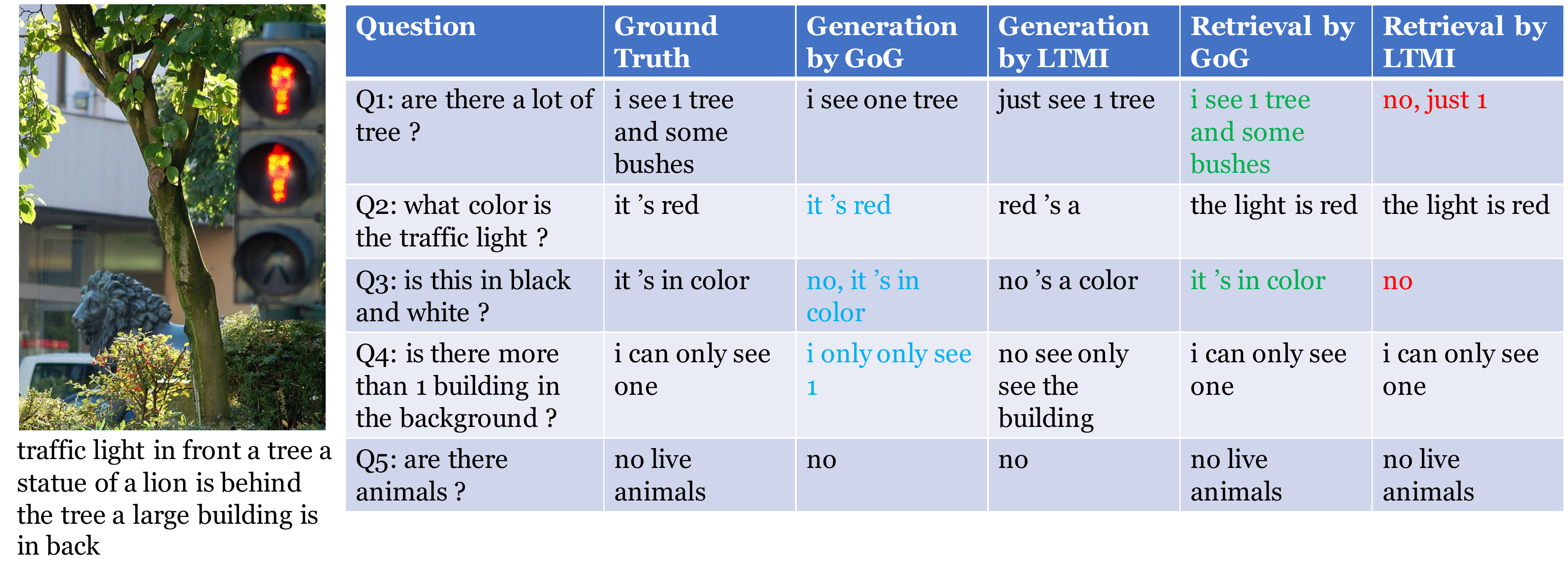}
  \end{overpic}
  } 
  \caption{Examples of dialogs generated and retrieved by our model and the LTMI baseline. Our model provides answers that are more accurate than LTMI (\textcolor{green}{green} denotes correct answers, and \textcolor{red}{red} denotes wrong answers). Results from our model are also more natural and comprehensive (highlighted in \textcolor{blue}{blue}).
  }\label{fig:dialog} 
\end{figure*}

\subsection{Human Study}
As shown in~\tabref{tab:humanstudy}, we conduct human study to further prove the effectiveness of our model. Our model achieves the highest scores both on the metric M1 and M2 compared with the previous model, LTMI~\cite{nguyenefficient}. These results show that our model can generate a better contextually and visually coherent response.

\begin{table}
  \centering
    \resizebox{0.70\columnwidth}!{
  \begin{tabular}{p{2.5cm}|c|c}
    \toprule
      & LTMI~\cite{nguyenefficient} & GoG\\
    \midrule
    Method 1 (M1) & 53 & 64\\
    \midrule
    Method 2 (M2) & 60  & 67 \\
    \bottomrule
  \end{tabular}
  }\vspace{3pt}
  \caption{Human evaluation on 100 sampled responses on VisDial val v1.0.  M1: percentage of responses pass the Turing Test. M2: percentage of responses evaluated better than or equal to human responses. }\label{tab:humanstudy} \vspace{-5pt}
\end{table}


\subsection{Qualitative Results}
As shown in~\figref{fig:example}, we visualize the learned attention maps. For the image, the colorful region means higher attention weights. We draw the bounding boxes of the first three highest scores. For the question, the word which has the darker color has higher attention weights. For dialog history, the darker QA pairs have a higher coreference score with the question. \figref{fig:dialog} provides some dialog examples, as a comparison of the results between GoG and the state-of-the-art LTMI model in the supplementary material. We have two observations by analyzing a set of randomly selected examples. Firstly, GoG generally provides more accurate answers. Secondly, GoG tends to provide longer and more natural human-like answers. More examples can be referred to Appendix~{\ref{appendix:exp}}.


\section{Related Work}
\subsection{Visual dialog}
For the visual dialog task~\cite{das2017visual}, GNN-EM~\cite{zheng2019reasoning} utilizes an EM-style GNN to conduct the textual coreference, which regards the caption and the previous question-answer (QA) pairs as observed nodes, and the current answer is deemed as an unobserved node inferred using EM algorithms~\cite{moon1996expectation} on the textual contexts. FGA~\cite{schwartz2019factor} realizes a factor graph attention mechanism, which constructs the graph over all the multi-modal features and estimates their interactions. DualVD~\cite{jiang2020dualvd} constructs a scene graph to represent the image while embedding both relationships provided by~\cite{zhang2019large} and original object detection features~\cite{anderson2018bottom}. CAG~\cite{guo2020iterative} focuses on an iterative question-conditioned context-aware graph, including both fine-grained visual-objects and textual-history semantics. In this paper, we model explicit complex relations within and among visual content, dialog history and the current question and design a graph-over-graph structure which are different from graph-based models mentioned above.

\subsection{Graph Neural Network}
Graph neural networks~\cite{kipf2016semi,velivckovic2017graph,xinyi2018capsule,zhang2019heterogeneous} have attracted attention in various tasks~\cite{wang2019neighbourhood,liu2018structure,gu2019scene}. The core idea is to combine the graphical structural representation with neural networks, which is suitable for reasoning-style tasks. For visual question answering, Liu et al.~\cite{teney2017graph} propose the first GNN-based approach, which builds a scene graph of the image and parses the sentence structure of the question, and calculates their similarity weights. Li et al.~\cite{li2019relation} propose to encode each image into a graph and model multi-type inter-object relations via a graph attention mechanism, such as spatial relations and semantic, and implicit relations~\cite{li2019relation}. Huang et al.~\cite{huang2020aligned} propose a novel dual-channel graph convolutional network for better combining visual and textual advantages. These approaches are limited to built independent graphs. There is no exploration of the coreference among dialog history and relations between graphs in the approach mentioned above.

\section{Conclusion}
In this paper, we present a relation-aware graph-over-graph network (GoG), a novel framework for visual dialog, which models and reasons the explicit complex relations among visual content, dialog history, and the current question. GoG exploits the graph-over-graph structure to obtain three relation-aware multi-modal representation which can be added to prior visual dialog models. Experimental results on two large-scale datasets show that our approach improves the previous models by a significant margin.

\bibliographystyle{acl_natbib}
\bibliography{anthology,acl2021}

\appendix
\section{Relation-aware Graph-over-Graph Network}\label{appendix:model}
\subsection{Question Graph Construction}
In practice, we observe that two words in a sentence usually hold certain relations. Such relations can be identified by the Neural Dependency Parsing~\cite{dozat2017deep}.  As shown in~\tabref{tab:dependency}, we list a part of commonly-used dependency relations.

\subsection{Attention Alignment Module}
After obtaining relation-aware features, we fuse the question representation $q^{*}$, history representation $h^{*}$, visual representation $v^{*}$ through a multi-modal fusion strategy. We can use any existing multi-modal fusion method to learn a joint representation $\mathcal{J}$:
\begin{equation}
    \mathcal{J} = \mathcal{F}(q^{*},h^{*},v^{*};\Theta),
\end{equation}
where $\mathcal{J}$ is a multi-modal fusion method and $\Theta$ are trainable parameters of the fusion module. Here we utilize an efficient attention mechanism method~\cite{nguyenefficient} to fuse the multi-modal information, which is the state-of-the-art model in visual dialog.

Let $A_X(Y)$ denotes the efficient attention mechanism~\cite{nguyenefficient} from the information $X$ to the information $Y$. For example, $A_{v^{*}}(v^{*})$ denotes the efficient self-attention. The fused visual representation is obtained as follows:
\begin{eqnarray}
    v_{contact} &=& [A_{v^{*}}(v^{*}), A_{q^{*}}(v^{*}), A_{h^{*}}(v^{*})], \\
    \overline{v}^{\prime} &=& {\rm LayerNorm}(\sigma (v_{concat}W_{v^{*}}) + v^{*}), \\
    a_V &=& {\rm softmax}(W_{V_1}\sigma (W_{V_2}\overline{v}^{\prime})), \\
    \overline{v} &=& \sum_{i=1}^{\mu}a_{V,i}\overline{v}^{\prime}_i,
\end{eqnarray}
where $W_{v^{*}}$, $W_{V_1}$, $W_{V_2}$ are learned parameters. $\overline{q}$ and $\overline{h}$ can be obtained similarly. Thus, the joint representation $\mathcal{J}$ is obtained:
\begin{equation}
    \mathcal{J} = W_J[\overline{q}, \overline{h}, \overline{v}],
\end{equation}
where $W_J$ is a learned parameter.

\begin{table}
  \centering
    \resizebox{0.90\columnwidth}!{
  \begin{tabular}{ccc}
    \toprule
      Relations & Relation Description & Proportion\\
    \midrule
    nsubj & nominal subject & 16.1\% \\
    root & root node & 16.0\% \\
    dep & dependent & 15.7\% \\
    punct & punctuation & 14.3\% \\
    det & determiner & 9.0\% \\
    cop & copula & 9.0\% \\
    prep & prepositional modifier & 4.6\% \\
    aux & auxiliary & 4.0\% \\
    pobj & object of a preposition & 3.6\% \\
    amod & adjective modifier & 3.2\% \\
    advmod & adverbial modifier & 2.5\% \\
    dobj & direct object & 1.5\% \\
    \bottomrule
  \end{tabular}
  }\vspace{3pt}
  \caption{The main categories of relations classified by the dependency parsing tool~\cite{dozat2017deep} in VisDial v1.0 training split~\cite{das2017visual}. }\label{tab:dependency}
\end{table}

\subsection{Generative and Discriminative Decoders}
Following Das et al.~\cite{das2017visual}, we consider both generative and discriminative decoders to score the candidate answers using the likelihood scores and the log-likelihood scores, respectively.

\begin{table*}
\centering
\resizebox{0.95\textwidth}!{
\begin{tabular}{lccccc|cccccc}
\toprule
\multirow{2}{*}{Model} &
\multicolumn{5}{c}{VisDial v0.9 (val)} &
\multicolumn{6}{c}{VisDial v1.0 (val)} \\
\cline{2-6}   \cline{7-12}
& MRR $\uparrow$ & R@1 $\uparrow$ & R@5 $\uparrow$ & R@10 $\uparrow$ & Mean $\downarrow$ & NDCG $\uparrow$ & MRR $\uparrow$ & R@1 $\uparrow$ & R@5 $\uparrow$ & R@10 $\uparrow$ & Mean $\downarrow$  \\
\midrule
\multicolumn{12}{c}{Fusion-based Models}\\
\midrule
LF~\cite{das2017visual} & 51.99 & 41.83 & 61.78 & 67.59 & 17.07
& - & - & - & - & - & -\\
HRE~\cite{das2017visual} & 52.37	& 42.23 & 62.28	& 68.11	& 16.97 
& - & - & - & - & - & -\\
\midrule
\multicolumn{12}{c}{Attention-based Model} \\
\midrule
MN~\cite{das2017visual} & 52.59 & 42.29 & 62.85 & 68.88 & 17.06
& 51.86 & 47.99 & 38.18 & 57.54 & 64.32 & 18.60\\
HCIAE~\cite{lu2017best} & 53.86 & 44.06 & 63.55 & 69.24 & 16.01 
& 59.70 & 49.07 & 39.72 & 58.23 & 64.73 & 18.43 \\
CorefNMN~\cite{Kottur2018VisualCR} & 53.50 & 43.66 & 63.54 & 69.93 & 15.69 
& - & - & - & - & - & - \\
CoAtt~\cite{wu2018you} & 54.11 & 44.32 & 63.82 & 69.75 & 16.47 
& 59.24 & 49.64 & 40.09 & 59.37 & 65.92 & 17.86 \\
RvA~\cite{niu2019recursive} & 55.43 & 45.37 & 65.27 & \underline{72.97} & \underline{10.71}
& - & - & - & - & - & - \\
DVAN~\cite{guo2019dual} & 55.94 & \underline{46.58} & 65.50 & 71.25 & 14.79
& - & - & - & - & - & - \\
Primary~\cite{guo2019image} & - & - & - & - & -
& - & 49.01 & 38.54 & 59.82 & 66.94 & 16.60 \\
ReDAN~\cite{gan2019multi} & - & - & - & - & -
& 60.47 & 50.02 & 40.27 & 59.93 & 66.78 & 17.40 \\
DMRM~\cite{chen2020dmrm} & \underline{55.96} & 46.20 & \underline{66.02} & 72.43 & 13.15
& - & 50.16 & 40.15 & 60.02 & 67.21 & 15.19 \\
DAM~\cite{jiang2020dam} & - & - & - & - & -
& 60.93 & \underline{50.51} & \underline{40.53} & \underline{60.84} & 67.94 & 16.65 \\
\midrule
\multicolumn{12}{c}{Pretraining-based Model} \\
\midrule
VDBERT~\cite{wang2020vd} & 55.95 & 46.83 & 65.43 & 72.05 & 13.18
& - & - & - & - & - & - \\
\midrule
\multicolumn{12}{c}{Graph-based Model} \\
\midrule
KBGN~\cite{jiang2020kbgn} & - & - & - & - & -
& 60.42 & 50.05 & 40.40 & 60.11 & 66.82 & 17.54 \\
\midrule
LTMI~\cite{nguyenefficient}$^\dagger$ & 55.85 & 46.07 & 65.97 & 72.44 & 14.17
& \underline{61.61} & 50.38 & 40.30 & 60.72 & \underline{68.44} & \underline{15.73} \\
GoG-Gen (Ours) & \bf{56.32} & \bf{46.65} & \bf{66.41} & 72.69 & 13.78
& \bf{62.63} & \bf{51.32} & \bf{41.25} & \bf{61.83} & \bf{69.44} & \bf{15.32} \\
GoG-Multi-Gen (Ours) & 56.89 & 47.04 & 66.92 & 72.87 & 13.45
& 63.35 & 51.80 & 41.78 & 62.23 & 69.79 & 15.16 \\
GoG-Multi (Ours) & 59.38& 48.58 & 71.33 & 78.78 & 9.94
& 65.20 & 55.38 & 43.93 & 68.22 & 76.75 & 9.98 \\
\bottomrule
\end{tabular}}
\vspace{3pt}
\caption{Main comparisons on both VisDial v0.9 and v1.0 datasets using the generative decoder. $\dagger$ denotes that we re-implemented the model. Underline indicates the highest performance among previous approaches except pretraining-base models.}
\label{tab:gen_appendix}
\end{table*}

\paragraph{Generative Decoder}
Following Das et al.~\cite{das2017visual}, we design the generative decoder to score the candidate answers using the log-likelihood scores. Specifically, the generative decoder utilizes a two-layer LSTM~\cite{hochreiter1997long} to generate an answer using the context vector $\mathcal{J}$ as the initial hidden state. In the training phase, the generative decoder generates the next token based on the current token from the ground truth answer. In detail, we first append the special token ``SOS'' at the beginning of the ground truth answer and ``EOS'' at the end. We use Glove~\cite{pennington2014glove} to initialize the embedding and obtain the embedding vectors $a_{gt} = [w_0, w_1, \dots, w_N]$ where $w_0$ is the embedding of ``SOS'' and $w_N$ is the embedding of ``EOS''. The hidden state $h_n$ at timestep $n$ is computed as follows:
\begin{equation}
    h_n = {\rm LSTM}(w_{n-1}, h_{n-1}),
\end{equation}
where $h_0$ is intializaed by $\mathcal{J}$. Then we obtain the log-likelihood of $n$-th word as follows:
\begin{equation}
    p = {\rm logsoftmax}(Wh + b),
\end{equation}
where $W$ and $b$ are learned parameters. In the training phase, we minimize the summation of the negative log-likelihood $\mathcal{L_G}$ defined by:
\begin{equation}
    \mathcal{L_G} = -\sum_{n=1}^Np_n.
\end{equation}
In the validation and test phase, we compute the summation $s_i$ of the log-likelihood for each candidate answer $\hat{a}_i$:
\begin{equation}
    s_i = \sum_{n=1}^Np_n^{\hat{a}_i}.
\end{equation}
Then, the rankings of the candidate answers are derived as ${\rm softmax}(s_1, \dots, s_{100})$.

\paragraph{Discriminative Decoder}
A discriminative decoder outputs the likelihood score for each of 100 candidate answers for the current question. Similar to the generative decoder, we use LSTM to obtain the hidden state $h_n$ for $b$-th word but we do not use context vector $J$ to initialize the $h_0$. The representation of each candidate answer is $a_i = h_N$. The score $p_i$ for $i$-th candidate answer is computed by:
\begin{equation}
    p = {\rm logsoftmax}(a_1^T\mathcal{J}, \dots, a_{100}^T\mathcal{J})
\end{equation}
In the test phase, we sort the candidate answers using these scores. In the training phase, the cross-entropy loss $\mathcal{L_D}$.

\subsection{Multi-Task Learning}
According to \cite{nguyenefficient}, we apply our GoG to the state-of-the-art model~\cite{nguyenefficient} in the multi-task learning setting that accuracy is improved by training the entire network using the two decoders simultaneously. This is simply done by minimizing the sum of the losses, $\mathcal{L_D}$ for the discriminative one and $\mathcal{L_G}$ for the generative one:
\begin{equation}
    \mathcal{L} = \mathcal{L_D} + \mathcal{L_G}
\end{equation}
The increase in performance may be attributable to the synergy of learning two tasks while sharing the same encoder.

\section{Experiments}\label{appendix:exp}

\subsection{Main Results}
\paragraph{Comparison with previous approaches using generative decoders.}
As shown in~\tabref{tab:gen_appendix}, we provide the full comparison with all the previous generative approaches.

\begin{figure*} 
\centering
\scalebox{0.90}{
  \begin{overpic}[width=\textwidth]{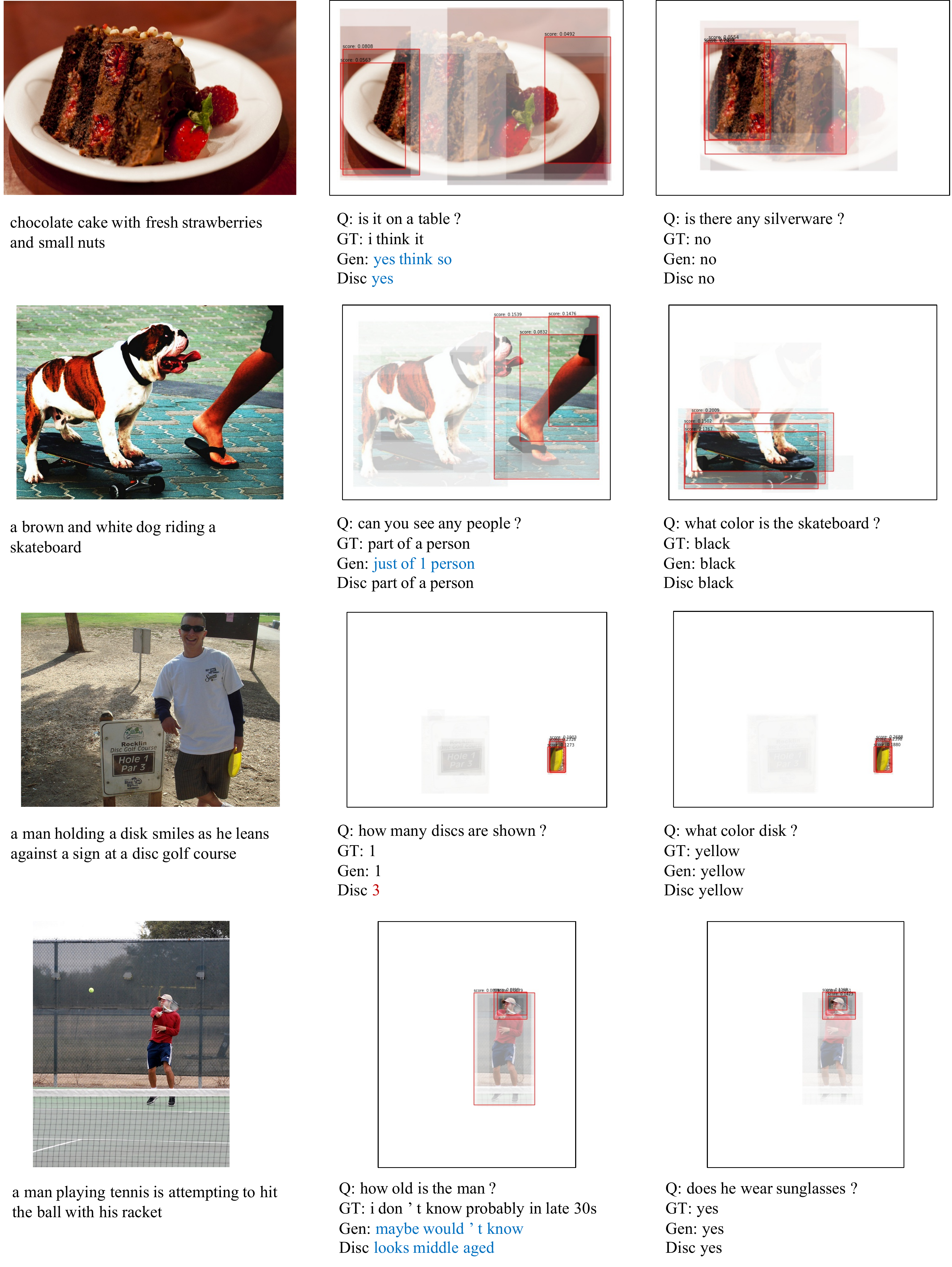}
  \end{overpic}
  }\vspace{-5pt}
  \caption{Examples of dialogs generated and retrieved by our model. \textcolor{blue}{blue} denotes has the same meaning with the ground truth and \textcolor{red}{red} denotes wrong answers.
  }\label{fig:case1}
\end{figure*}

\subsection{Qualitative Results}
More examples generated and retrieved by our GoG are provided in~\figref{fig:case1}. Due to the limited number of pages, we only provide an additional example of~\figref{fig:case1}.



\end{document}